\documentclass[10pt,twocolumn,letterpaper]{article}

\usepackage{cvpr}
\usepackage{times}
\usepackage{epsfig}
\usepackage{graphicx}
\usepackage{amsmath}
\usepackage{amssymb}
\usepackage{subfigure}
\usepackage{multirow}
\usepackage{bm}
\usepackage{url}
\usepackage{float} 
\usepackage{pdfpages}

\usepackage[pagebackref=true,breaklinks=true,letterpaper=true,colorlinks,bookmarks=false]{hyperref}

\cvprfinalcopy % *** Uncomment this line for the final submission

\def\cvprPaperID{****} % *** Enter the CVPR Paper ID here

\begin{document}

%%%%%%%%% TITLE
\title{Two Causal Principles for Improving Visual Dialog}

\author{Jiaxin Qi$^{1}$
~~~~~~~Yulei Niu$^{2}$~~~~~~~Jianqiang Huang$^{1,3}$\protect\thanks{Corresponding author.}~~~~~~~Hanwang Zhang$^{1}$\\
$^{1}$Nanyang Technological University,
    $^{2}$Renmin University of China,\\
    $^{3}$Damo Academy, Alibaba Group,\\
{\tt\small jiaxin003@e.ntu.edu.sg,
niu@ruc.edu.cn,
jianqiang.jqh@gmail.com,
hanwangzhang@ntu.edu.sg}
}

\maketitle
%%%%%%%%% 0 ABSTRACT
\begin{abstract}
This paper unravels the design tricks adopted by us --- the champion team MReaL-BDAI --- for Visual Dialog Challenge 2019: \textbf{two causal principles} for improving Visual Dialog (VisDial). By ``improving'', we mean that they can promote almost every existing VisDial model to the state-of-the-art performance on the leader-board. Such a major improvement is only due to our careful inspection on the \textbf{causality} behind the model and data, finding that the community has overlooked two causalities in VisDial. Intuitively, \textbf{Principle 1} suggests: we should remove the direct input of the dialog history to the answer model, otherwise a harmful shortcut bias will be introduced; \textbf{Principle 2} says: there is an unobserved confounder for history, question, and answer, leading to spurious correlations from training data. In particular, to remove the confounder suggested in Principle 2, we propose several \textbf{causal intervention} algorithms, which make the training fundamentally different from the traditional likelihood estimation. Note that the two principles are \emph{model-agnostic}, so they are applicable in any VisDial model. The code is available at \url{https://github.com/simpleshinobu/visdial-principles}.\let\thefootnote\relax
\end{abstract}

%%%%%%%%% 1 BODY TEXT
\section{Introduction}
\label{section:1}
Given an image $I$, a dialog history of past Q\&A pairs: $H = \{(Q_1, A_1), ..., (Q_{t-1}, A_{t-1})\}$, and the current $t$-th round question $Q$, a Visual Dialog (VisDial) agent~\cite{das2017visual} is expected to provide a good answer $A$. Our community has always considered VQA~\cite{antol2015vqa} and VisDial as sister tasks due to their similar settings: Q\&A grounded by $I$ (VQA) and Q\&A grounded by $(I,H)$ (VisDial). Indeed, from a technical point view --- just like the VQA models --- a typical VisDial model first uses \emph{encoder} to represent $I$, $H$, and $Q$ as vectors, and then feed them into \emph{decoder} for answer $A$. Thanks to the recent advances in encoder-decoder frameworks in VQA~\cite{lu2016hierarchical,teney2018tips} and natural language processing~\cite{vaswani2017attention}, the performance (NDCG~\cite{visdial2019}) of VisDial in literature is significantly improved from the baseline 51.63\%~\cite{visdial2019challenge} to the state-of-the-art 64.47\%~\cite{DBLP:conf/acl/GanCKLLG19}.

\begin{figure}[!t]
\centering
\includegraphics[width=3.2in]{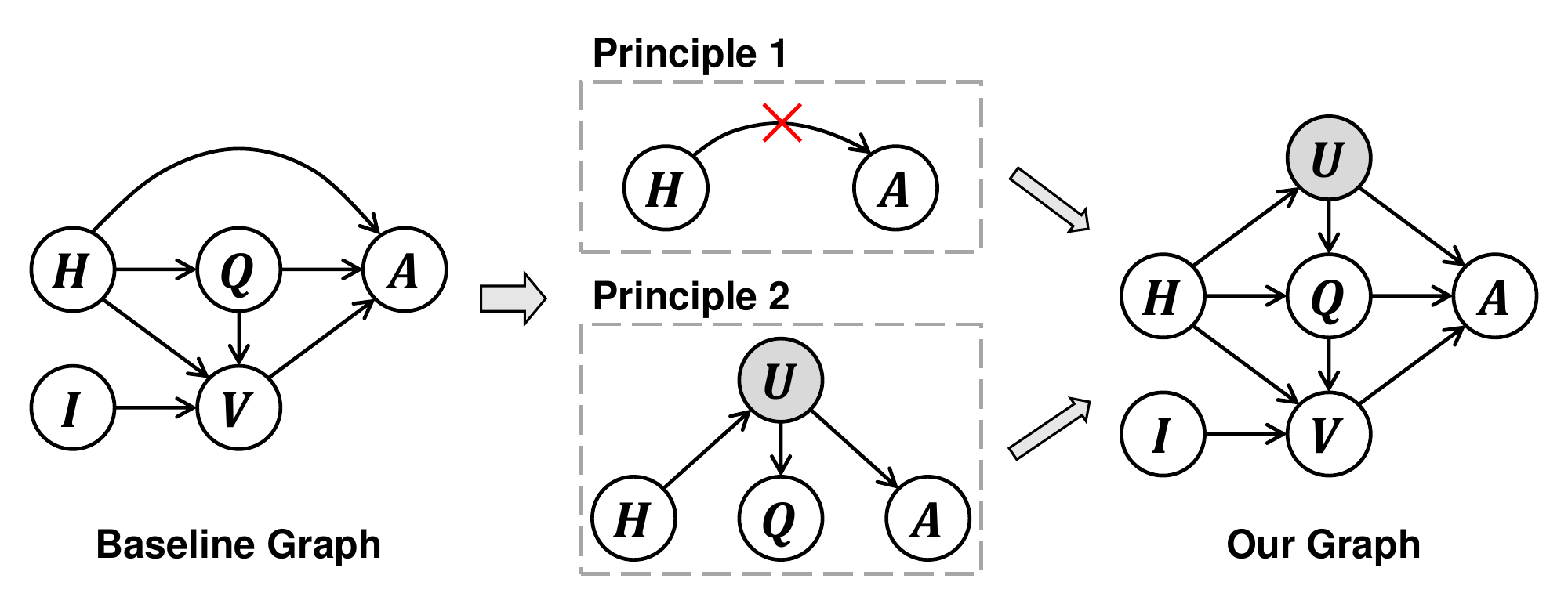}
\caption{Causal graphs of VisDial models (baseline and ours). $H$: dialog history. $I$: image. $Q$: question. $V$: visual knowledge. $A$: answer. $U$: user preference. Shaded $U$ denotes unobserved confounder. See Section~\ref{subsection:3_2} for detailed definitions.}
\label{fig:1}
\end{figure}

However, in this paper, we want to highlight an important fact: \emph{VisDial is essentially NOT VQA with history!} And this fact is so profound that all the common heuristics in the vision-language community --- such as the multimodal fusion~\cite{teney2018tips,yu2017multi} and attention variants~\cite{lu2016hierarchical,nam2017dual,niu2019recursive} --- cannot appreciate the difference. Instead, we introduce the use of \emph{causal inference}~\cite{pearl2009causal,pearl2016causal}: a graphical framework that stands in the cause-effect \emph{interpretation} of the data, but not merely the statistical \emph{association} of them. Before we delve into the details, we would like to present the main contributions: two causal principles, rooted from the analysis of the difference between VisDial and VQA, which lead to a performance leap --- a farewell to the 60\%-s and an embrace for the 70\%-s --- for all the baseline VisDial models\footnote{Only those with codes\&reproducible results due to resource limit.} in literature~\cite{das2017visual,lu2017best,wu2018you,niu2019recursive}, promoting them to the state-of-the-art in Visual Dialog Challenge 2019~\cite{visdial2019challenge}. 
\newtheorem{principle}{Principle}
\begin{principle}
(\textbf{P1}): Delete link $H\rightarrow A$.
\end{principle}
\begin{principle}
(\textbf{P2}): Add one new (unobserved) node $U$ and three new links: $U\leftarrow H$, $U\rightarrow Q$, and $U\rightarrow A$.
\end{principle}
Figure~\ref{fig:1} compares the causal graphs of existing VisDial models and the one applied with the proposed two principles. Although a formal introduction of them is given in Section~\ref{subsection:3_2}, now you can simply understand the nodes as data types and the directed links as modal transformations. For example, $V\rightarrow A$ and $Q\rightarrow A$ indicate that answer $A$ is the \emph{effect} \emph{caused} by visual knowledge $V$ and question $Q$, through a transformation, \eg, a multi-modal encoder.  

\begin{figure}[!t]
\centering
\subfigure[A Typical $H\rightarrow A$ Bias]{
\label{fig:2_a}
\includegraphics[width=3.2in]{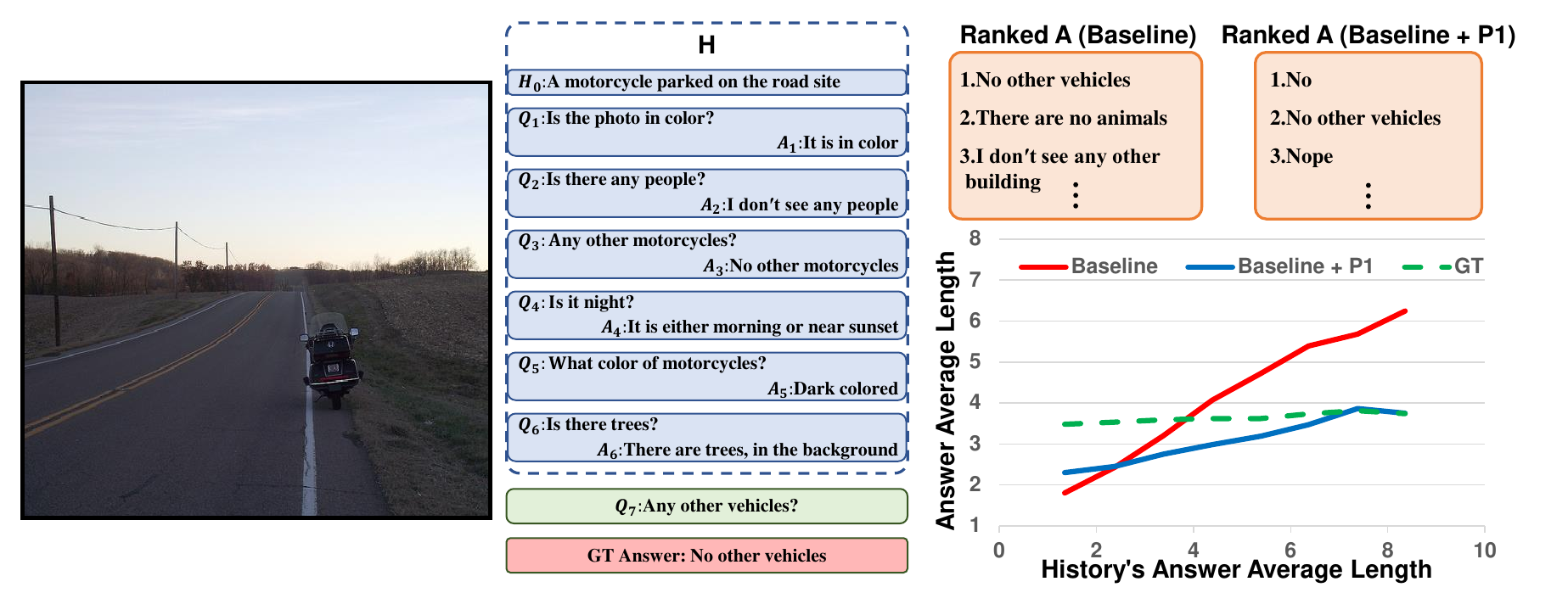}
}
\subfigure[User Preference]{
\label{fig:2_b}
\includegraphics[width=3.2in]{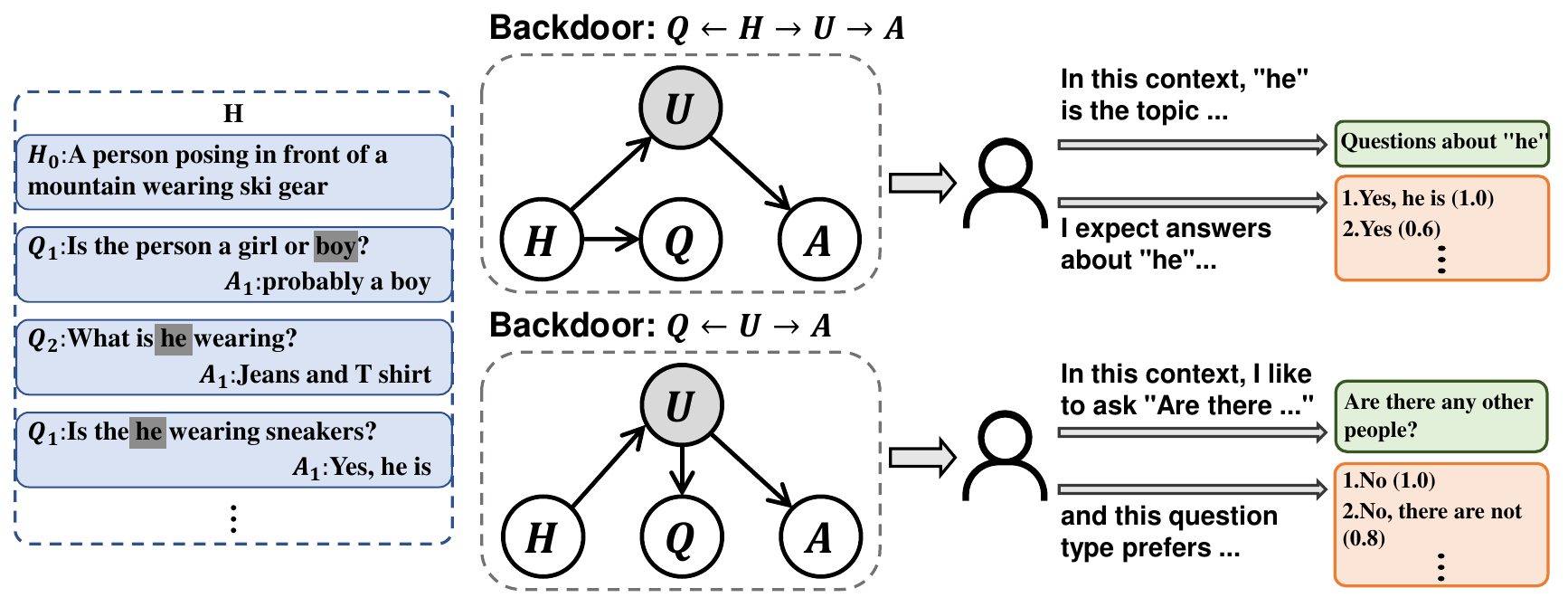}
}
\caption{The illustrative motivations of the two causal principles: (a) P1 and (b) P2. }
\label{fig:2}
\vspace{-5pt}
\end{figure}

P1 suggests that we should remove the \emph{direct} input of dialog history to the answer model. This principle contradicts most of the prevailing VisDial models~\cite{das2017visual,jain2018two,wu2018you,niu2019recursive,yang2019making,DBLP:conf/emnlp/KangLZ19,DBLP:conf/acl/GanCKLLG19,schwartz2019factor}, which are based on the widely accepted intuition: the more features you input, the more effective the model is. It is mostly correct, but only with our discretion of the data generation process. In fact, the VisDial~\cite{das2017visual} annotators were not allowed to copy from the previous Q\&A, \ie, $H\nrightarrow A$, but were encouraged to ask consecutive questions including co-referenced pronouns like ``it'' and ``those'', \ie, $H\rightarrow Q$, and thus the answer $A$ is expected to be only based on question $Q$ and reasoned visual knowledge $V$. Therefore, a good model should reason over the context $(I, H)$ with $Q$ but not to memorize the bias. However, the direct path $H\rightarrow A$ will contaminate the expected causality. Figure~\ref{fig:2_a} shows a very ridiculous bias observed in all baselines without P1: the top answers are those with length closer to the average length in the history answers. We will offer more justifications for P1 in Section~\ref{subsection:4_1}.

P2 implies that the model training only based on the association between $(I, H, Q)$ and $A$ is spurious. By ``spurious'', we mean that the effect on $A$ caused by $(I, H, Q)$ --- the goal of VisDial --- is \emph{confounded} by an unobserved variable $U$, because it appears in every undesired causal path (\textit{a.k.a.}, backdoor~\cite{pearl2016causal}), which is an indirect causal link from input $(I, H, Q)$ to output $A$: $Q\leftarrow U\rightarrow A$ and $Q\leftarrow H\rightarrow U\rightarrow A$. We believe that such unobserved $U$ should be \emph{users} as the VisDial dataset essentially brings humans in the loop. Figure~\ref{fig:2_b} illustrates how the user's hidden preference confounds them. Therefore, during training, if we focus only on the conventional likelihood $P(A|I, H, Q)$, the model will inevitably be biased towards the spurious causality, \eg, it may score answer ``Yes, he is'' higher than ``Yes'', merely because the users prefer to see a ``he'' appeared in the answer, given the history context of ``he''. It is worth noting that the confounder $U$ is more impactful in VisDial than in VQA, because the former encourages the user to rank similar answers subjectively while the latter is more objective. A plausible explanation might be: VisDial is interactive in nature and a not quite correct answer is tolerable in one iteration (\ie, dense prediction); while VQA has only one chance, which demands accuracy (\ie, one-hot prediction).

By applying P1 and P2 to the baseline causal graph, we have the proposed one (the right one in Figure~\ref{fig:1}), which serves as a \emph{model-agnostic} roadmap for the causal inference of VisDial. To remove the spurious effect caused by $U$, we use the \emph{do-calculus}~\cite{pearl2016causal} $P\left(A|do(I,H,Q)\right)$, which is fundamentally different from the conventional likelihood $P(A|I, H, Q)$: the former is an active \emph{intervention}, which cuts off $U\rightarrow Q$ and $H\rightarrow Q$, and sample (where the name ``calculus'' is from) every possible $U|H$, seeking the true effect on $A$ only caused by $(I,H,Q)$; while the latter likelihood is a passive \emph{observation} that is affected by the existence of $U$. The formal introduction and details will be given in Section~\ref{subsection:4_3}. In particular, given the fact that once the dataset is ready, $U$ is no longer observed, we propose a series of effective approximations in Section~\ref{section:5}.  

We validate the effectiveness of P1 and P2 on the most recent VisDial v1.0 dataset. We show significant boosts (absolute NDCG) by applying them in 4 representative baseline models: LF~\cite{das2017visual} ($\uparrow$16.42\%), HCIAE~\cite{lu2017best} ($\uparrow$15.01\%), CoAtt~\cite{wu2018you} ($\uparrow$15.41\%), and RvA~\cite{niu2019recursive} ($\uparrow$16.14\%). Impressively, on the official test-std server, we use an ensemble model of the most simple baseline LF~\cite{das2017visual} to beat our 2019 winning performance by 0.2\%, a more complex ensemble to beat it by 0.9\%, and lead all the single-model baselines to the state-of-the-art performances.

%%%%%%%%% 2 RELATED WORK
\section{Related Work}
\label{section:2}
\noindent\textbf{Visual Dialog.} Visual Dialog~\cite{das2017visual,de2017guesswhat} is more interactive and challenging than most of the vision-language tasks, \eg, image captioning~\cite{yao2018exploring,yang2019auto,anderson2018bottom} and VQA~\cite{antol2015vqa,teney2018tips,tang2019learning,tang2020unbiased}. Specifically, Das~\etal~\cite{das2017visual} collected a large-scale free-form visual dialog dataset VisDial~\cite{buhrmester2011amazon}. They applied a novel protocol: during the live chat, the questioner cannot see the picture and asks open-ended questions, while the answerer gives free-form answers. Another dataset GuessWhat?! proposed by~\cite{de2017guesswhat} is a goal-driven visual dialog: questioner should locate an unknown object in a rich image scene by asking a sequence of closed-ended ``yes/no'' questions. We apply the first setting in this paper. Thus, the key difference is that the users played an important role in the data collection process.

All of the existing approaches in the VisDial task are based on the typical encoder-decoder framework~\cite{jain2018two,guo2019image,seo2017visual,DBLP:conf/acl/GanCKLLG19,schwartz2019factor,zheng2019reasoning}. They can be categorized by the usage of history. 1) Holistic: they treat history as a whole to feed into models like HACAN~\cite{yang2019making}, DAN~\cite{DBLP:conf/emnlp/KangLZ19} and CorefNMN~\cite{kottur2018visual}. 2) Hierarchical: they use a hierarchical structure to deal with history like HRE~\cite{das2017visual}. 3) Recursive: RvA~\cite{niu2019recursive} uses a recursive method to process history. However, they all overlook the fact that the history information should not be directly fed to the answer model (\ie, our proposed Principle 1). The baselines we used in this paper are LF~\cite{das2017visual}: the earliest model, HCIAE~\cite{lu2017best}: the first model to use history hierarchical attention, CoAtt~\cite{wu2018you}: the first one to a co-attention mechanism, and  RvA~\cite{niu2019recursive}: the first one for a tree-structured attention mechanism.

\noindent\textbf{Causal Inference.} Recently, some works~\cite{nair2019causal,bengio2019meta,mahajan2019preserving,singh2019biased,wang2020visual,yang2020deconfounded} introduced causal inference into machine learning, trying to endow models the abilities of pursuing the cause-effect. In particular, we use the Pearl's structural causal model (SCM) proposed by~\cite{pearl2016causal} to hypothesize the data generation process, which is a model-agnostic framework that reflects the nature of the data.

%%%%%%%%% 3 Encoder-Decoder as Causal Graph
\section{Visual Dialog in Causal Graph}
\label{section:3}
In this section, we formally introduce the visual dialog task and describe how the popular encoder-decoder framework follows the baseline causal graph shown in Figure~\ref{fig:1}. More details of causal graph can be found in~\cite{pearl2016causal,Judea2018thebookofwhy}.
\subsection{Visual Dialog Settings}
\label{subsection:3_1}
\noindent\textbf{Settings.} According to the definition of VisDial task proposed by Das~\etal~\cite{das2017visual}, at each time $t$, given input image $I$, current question $Q_t$, dialog history $H=\{C,(Q_1,A_1),\dots,(Q_{t-1},A_{t-1})\}$, where $C$ is the image caption, $(Q_i,A_i)$ is the $i$-th round Q\&A pair, and a list of 100 candidate answers $A_t = \{A_t^{(1)},\dots,A_t^{(100)}\}$. A VisDial model is evaluated by ranking candidate answers $A_t$.

\noindent\textbf{Evaluation.} Recently, the ranking metric Normalized Discounted Cumulative Gain (NDCG) is adopted by the VisDial community~\cite{visdial2019}. It is different from the classification metric (\eg, top-1 accuracy) used in VQA. It is more compatible with the relevance scores of the answer candidates in VisDial rated by humans. NDCG requires to rank relevant candidates in higher places, rather than just to select the ground-truth answer.
\subsection{Encoder-Decoder as Causal Graph}
\label{subsection:3_2}
We first give the definition of causal graph, then revisit the encoder-decoder framework in existing methods using the elements from the baseline graph in Figure~\ref{fig:1}.

\noindent\textbf{Causal Graph}. Causal graph~\cite{pearl2016causal}, as shown in Figure~\ref{fig:1}, describes how variables interact with each other, expressed by a directed acyclic graph $\mathcal{G}=\{\mathcal{N},\mathcal{E}\}$ consisting of nodes $\mathcal{N}$ and directed edges $\mathcal{E}$ (\ie, arrows). $\mathcal{N}$ denote variables, and $\mathcal{E}$ (arrows) denote the causal relationships between two nodes, 
\ie, $A\rightarrow B$ denotes that $A$ is the cause and $B$ is the effect, meaning the outcome of $B$ is caused by $A$. Causal graph is a highly general roadmap specifying the causal dependencies among variables. 

As we will discuss in the following part, all of the existing methods can be revisited in the view of the baseline graph shown in Figure~\ref{fig:1}.

\noindent\textbf{Feature Representation and Attention in Encoder}. Visual feature is denoted as node $I$ in the baseline graph, which is usually a fixed feature extracted by Faster-RCNN~\cite{ren2015faster} based on ResNet  backbone~\cite{he2016deep} pre-trained on Visual Genome~\cite{krishna2017visual}. For language feature, the encoder firstly embeds sentence into word vectors, followed by passing the RNN~\cite{hochreiter1997long, DBLP:conf/ssst/ChoMBB14} to generate features of question and history, which are denoted as $\{Q,H\}$.

Most of existing methods apply attention mechanism~\cite{xu2015show} in encoder-decoder to explore the latent weights for a set of features. A basic attention operation can be represented as $\bm{\tilde{x}} = Att(\mathcal{X}, \mathcal{K})$ where $\mathcal{X}$ is the set of features need to attend, $\mathcal{K}$ is the key (\ie, guidance) and $\bm{\tilde{x}}$ is the attended feature of $\mathcal{X}$. Details can be found in most visual dialog methods~\cite{lu2017best,wu2018you,yang2019making}. In the baseline graph, the sub-graph $\{I \rightarrow V, Q \rightarrow V, H \rightarrow Q \rightarrow V\}$ denotes a series of attention operations for visual knowledge $V$. Note that the implementation of the arrows are not necessarily independent, such as co-attention~\cite{wu2018you}, and the process can be written as ${\textit{Input}}:\{I, Q, H\} \Rightarrow {\textit{Output}}:\{V\}$, where possible intermediate variables can be added as mediator nodes into the original arrows. However, without loss of generality, these mediators do not affect the causalities in the graph.

\noindent\textbf{Response Generation in Decoder}. After obtaining the features from the encoder, existing methods will fuse them and feed the fused ones into a decoder to generate an answer.
In the baseline graph, node $A$ denotes the answer model that decodes the fused features from $\{H\! \rightarrow \!A, Q\! \rightarrow \!A, V\! \rightarrow\! A\}$ and then transforms them into an answer sentence.
In particular, the decoder can be generative, \ie, to generate an answer sentence using RNN; or discriminative, \ie, to select an answer sentence by using candidate answer classifiers.

%%%%%%%%% 4 Principles
\section{Two Causal Principles}
\label{section:4}
\subsection{Principle 1}
\label{subsection:4_1}
When should we draw an arrow from one node pointing to another? According to the definition in Section~\ref{subsection:3_2}, the criterion is that if the node is the cause and the other one is the effect. Intrigued, let's understand P1 by discussing the rationale behind the ``double-blind'' review policy. Given three variables: ``Well-known Researcher'' ($R$), ``High-quality Paper'' ($P$), and ``Accept'' ($A$). From our community common sense, we know that $R\!\rightarrow\!P$ since top researchers usually lead high-quality research, and $P\rightarrow A$ is even more obvious. Therefore, for the good of the community, the double-blind prohibits the direct link $R\rightarrow A$ by author anonymity, otherwise the bias such as personal emotions and politics from $R$ may affect the outcome of $A$. 

The story is similar in VisDial. Without loss of generality, we only analyze the path $H\! \rightarrow\! Q\! \rightarrow A$. If we inspect the role of $H$, we can find that it is to help $Q$ resolve some co-references like ``it'' and ``their''. As a result, $Q$ listens to $H$. Then, we use $Q$ to obtain $A$. Here, $Q$ becomes a mediator which cuts off the direct association between $H$ and $A$ that makes $P(A| Q, H)\!=\!P(A| Q)$, like the ``High-quality Paper'' that we mentioned in the previous story. However, if we set an arrow from $H$ to $A$: $H\rightarrow A$, the undesirable bias of $H$ will be learned for the prediction of $A$, that hampers the natural process of VisDial, such as the interesting bias illustrated in Figure~\ref{fig:2_a}. Another example is discussed in Figure~\ref{fig:4} that $A$ prefers to match the words in $H$, even though they are literally nonsense about $Q$ if we add the direct link $H\rightarrow A$. After we apply P1, these phenomena will be relieved, such as the blue line illustrated in Figure~\ref{fig:2_a}, which is closer to the NDCG ground truth average  answer length , denoted as the green dashed line. Please refer to other qualitative studies in Section~\ref{subsection:6_5}.

\subsection{Principle 2}
\label{subsection:4_2}
\begin{figure}[t]
\centering
{
\subfigure[Confounder $U$]{
\label{fig:3_a}
\includegraphics[width=1in]{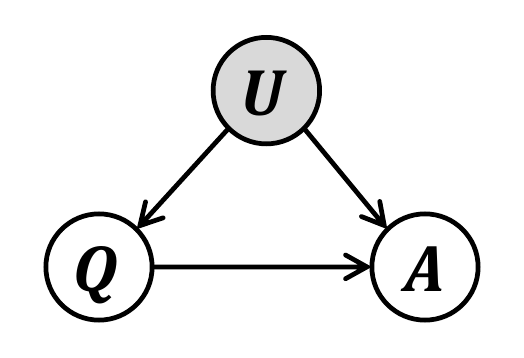}
}
\subfigure[Spurious Relation]{
\label{fig:3_f}
\includegraphics[width=1in]{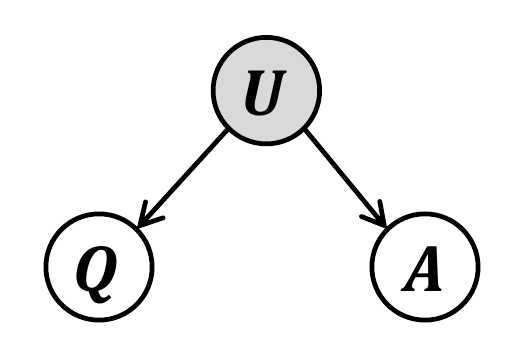}
}
\subfigure[\textit{do}-operator]{
\label{fig:3_b}
\includegraphics[width=1in]{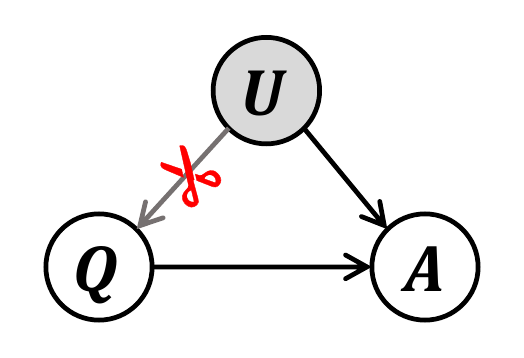}
}
\subfigure[Question Type]{
\label{fig:3_c}
\includegraphics[width=1in]{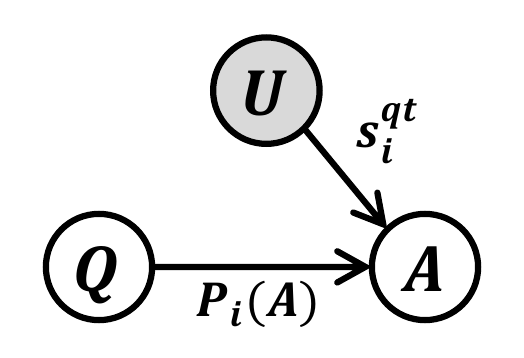}
}
\subfigure[Score Sampling]{
\label{fig:3_d}
\includegraphics[width=1in]{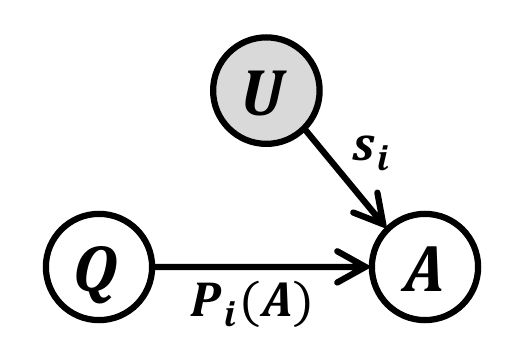}
}
\subfigure[Hidden Dictionary]{
\label{fig:3_e}
\includegraphics[width=1in]{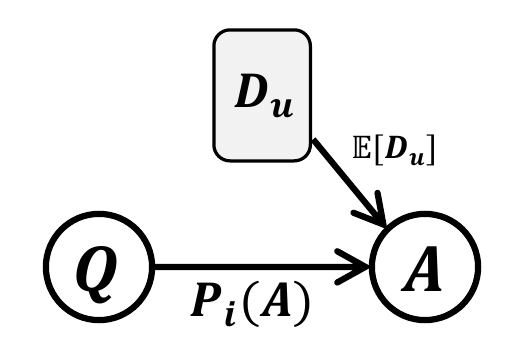}
}
}
\vspace{2pt}
\caption{Example of confounder, \textit{do}-operator and sketch causal graphs of our three attempts to de-confounder}
\vspace{-2pt}
\label{fig:3}
\end{figure}

Before discussing P2, we first introduce an important concept in causal inference~\cite{pearl2016causal}. In causal graph, the fork-like pattern in Figure~\ref{fig:3_a} contains a \emph{confounder} $U$, which is the common cause for $Q$ and $A$ (\ie, $Q \leftarrow\! U\! \rightarrow\! A$). The confounder $U$ opens a \emph{backdoor} path started from $Q$, making $Q$ and $A$ spuriously correlated even if there is no direct causality between them.

In the data generation process of VisDial, we know that not only both of the questioner and answerer can see the dialog history, but also the answer annotators can look at the history when annotating the answer. Their preference after seeing the history can be understood as a part of the human nature or subtleties conditional on a dialog context, and thus it has a causal effect on both $Q$ and $A$. Moreover, due to the fact that the preference is nuanced and uncontrollable, we consider it as an \emph{unobserved} confounder for $Q$ and $A$.

It is worth noting that the confounder hinders us to find the true causal effect. Let's take the graph in Figure~\ref{fig:3_f} as an example. The causal effect from $Q$ to $A$ is 0; however, we can quickly see that $P(A|Q)-P(A)$ is nonzero because $Q$ and $A$ are both influenced by $U$ and thus are correlated (thanks to Reichenbach's common cause principle~\cite{pearl2016causal}). That is, if we are given $Q$, the any likelihood change for $A$ will be sensible compared to nothing is given. Therefore, if we consider $P(A|Q)$ as our VisDial model, it will still predict nonsense answers even if $Q$ has nothing to do with $A$. As illustrated in Figure~\ref{fig:2_b}, model will prefer the candidates about ``he'' even though $Q$ is not given, that means it captures the confounder $U$ but not the true rationale between $Q$ and $A$. Next, we will introduce a powerful technique that makes the $Q$ and $A$ in Figure~\ref{fig:3_f} ``independent'', \ie, no causal relation.

\subsection{\textit{do}-calculus.}
\label{subsection:4_3}
The technique is \textit{do}-calculus introduced in~\cite{pearl2016causal,Judea2018thebookofwhy}. Specifically, \textit{do}$(Q=q)$ denotes that we deliberately assign a value $q$ to variable $Q$ (\ie, intervention), rather than passively observe $Q=q$. As illustrated in Figure~\ref{fig:3_b}, \textit{do}$(Q=q)$ can be understood as cutting all the original incoming arrows to $Q$, and then making $Q$ and $U$ independent.  Therefore, we can have the well-known backdoor adjustment~\cite{pearl2016causal}: $P(A|do(Q=q))=\sum\nolimits_u P(A|Q = q, u)P(u)$. Note that this is different from Bayes rule $P(A|Q=q)=\sum\nolimits_u P(A|Q = q, u)P(u|Q=q)$ thanks to the independence $P(u|Q=q) = P(u)$ introduced by  \textit{do}-calculus. Let's revisit Figure~\ref{fig:3_f} by using \textit{do}-calculus. We can find that $P(A|do(Q=q)) - P(A) = 0$, that is to say, any intervention of $Q$ will not influence the probability of $A$, meaning the correct relation between $Q$ and $A$: no causal relation. Therefore, $P(A|do(Q=q))$ should be the objective answer model in VisDial.

For our proposed graph of VisDial shown in Figure~\ref{fig:1}, we can use intervention \emph{do}$(Q,H,I)$ and the backdoor adjustment to obtain our overall model. Here, we slightly abuse the notation \emph{do}$(Q,H,I)$ as \emph{do}$(Q=q,H=h,I=i)$:
\begin{equation}
\centering
\label{eq:2}
\begin{split}
& P(A|\textit{do}(Q,H,I))\\
& = \sum\nolimits_u P(A|\textit{do}(Q,H,I), u)P(u|\textit{do}(Q,H,I))\\
& = \sum\nolimits_u P(A|\textit{do}(Q), H, I, u)P(u|H)\\
& = \sum\nolimits_u P(A|Q, H, I, u)P(u|H).
\end{split}
\end{equation}
The detailed derivation and proof can be found in supplementary materials. 

So far, we have provided all the ingredients of the baseline causal graph, two proposed principles and their theoretical solution: \textit{do}-calculus. Next, we will introduce some implementations for the proposed solution in Eq.~\eqref{eq:2}.
%%%%%%%%% 5 Improved Visual Dialog
\section{Improved Visual Dialog Models}
\label{section:5}
It is trivial to implement P1 and we will provide its training details in Section~\ref{subsection:6_3}. For P2, since $U$ is unobserved, it is impossible to sample $u$ in Eq.~\eqref{eq:2} directly. Therefore, our technical contribution is to introduce 3 approximations. For notation simplicity, we first re-write Eq.~\eqref{eq:2} as:

\begin{equation}
\setlength\abovedisplayskip{2pt}
\setlength\belowdisplayskip{2pt}
\centering
\label{eq:3}
P(A|\textit{do}(Q,H,I))=\sum\nolimits_u P_u(A)P(u|H),
\end{equation}
where $P_u(A) := P(A|Q,H, I, u)$. 
\subsection{Question Type} 
\label{subsection:5_1}
Since we cannot directly sample $u$ from the unobserved confounder, we use the $i$-th answer candidate $a_i$ as a delegate for sample $u$. That is because $a_i$ is a sentence observed from the ``mind'' of user $u$ during dataset collection. Then, $\sum\nolimits_uP_u(A)P(u|H)$ can be approximated as $\sum\nolimits_iP_i(A)P(a_i|H)$. We further use $p(a_i|QT)$ to approximate $P(a_i|H)$ because of two reasons: First, $P(a_i|H)$ essentially describes a prior knowledge about $a_i$ without comprehending the whole $\{Q,H,I\}$ triplet. A similar scenario is that if we know the QT (question type), \eg, ``what color'', the answer candidates denoting colors have higher probabilities without even comprehending the question details. Second, QT is extracted from question $Q$, which is a descendent of history $H$ in our graph, indicating that QT partially reveals $H$~\cite{pearl2016causal}. In practice, we manually define some question types, each of which has a certain answer frequency. For each dialog round, a normalized score $s_i^{qt}:=p(a_i|QT)$ (\ie., $\sum\nolimits_i s_i^{qt} =1$) of each candidate $a_i$ will be calculated according to the frequency of $a_i$ under question type $qt$. More details are given in Section~\ref{subsection:6_3}. Finally, we have the approximation for Eq.~\eqref{eq:3}:
\begin{equation}
\centering
\label{eq:4}
\sum\nolimits_u P_u(A)P(u|H) \approx \sum\nolimits_i P_i(A)\cdot s_i^{qt},
\end{equation}
where $P_i(A)=\text{softmax}(f_s(\bm{e}_i,\bm{m}))$, $f_s$ is a similarity function, $\bm{e}_i$ is the embedding of candidate $a_i$, $\bm{m}$ is the joint embedding for $\{Q,I,H\}$, and the sketch graph is shown in Figure~\ref{fig:3_c}. Since question type is observed from $Q$, the approximation $p(a_i|QT)$ undermines the prior assumption of the backdoor adjustment in Eq.~\eqref{eq:2} (\ie, the prior $p(u|H)$ cannot be conditional on $Q$). Fortunately, $QT$ is only a small part of $Q$ (\ie, the first few words) and thus the approximation is reasonable.

\subsection{Answer Score Sampling}
\label{subsection:5_2}

Since the question type implementation slightly undermines the backdoor adjustment, we will introduce a better approximation which directly samples from $u$: Answer Score Sampling. This implementation is also widely known as our previously proposed \textbf{dense fine-tune} in community~\cite{report2019}. 

We still use $a_i$ to approximate $u$, and we use the (normalized) ground-truth NDCG score $s_i$ annotated by the humans to approximate $P(a_i|H)$. Note that $s_i$ directly reveals human preference for $a_i$ in the context $H$ (\ie, the prior $P(a_i|H)$). In practice, we use the subset of training set with dense annotations to sample $s_i$. Therefore, we have:
\begin{equation}
\centering
\label{eq:5}
\sum\nolimits_u P_u(A)P(u|H) \approx \sum\nolimits_i P_i(A)\cdot s_i,
\end{equation}
and the sketch graph is illustrated in Figure~\ref{fig:3_d}. In practice, Eq.~\eqref{eq:5} can be implemented using different loss functions. Here we give three examples:

\noindent\textbf{Weighted Softmax Loss ($R_1$).} We extend the log-softmax loss as a weighted form, where $P_i(A)$ is denoted by $\log(\text{softmax}(p_i))$, $p_i$ denotes the logit of candidate $a_i$, and $s_i$ is corresponding normalized relevance score.

\noindent\textbf{Binary Sigmoid Loss ($R_2$).} This loss is close to the binary cross-entropy loss, where $P_i(A)$ represents $\log(\text{sigmoid}(p_i))$ or $\log(\text{sigmoid}(1-p_i))$, and $s_i$ represents corresponding normalized relevance score.

\noindent\textbf{Generalized Ranking Loss ($R_3$).} Note that the answer generation process can be viewed as a ranking problem. Therefore, we derive a ranking loss that $P_i(A)$ is $\log\frac{\exp(p_i)}{\exp(p_i) + \sum_{j\in G}\exp(p_j)}$, where $G$ is a group of candidates which have lower relevance scores than candidate $a_i$ and $s_i$ is normalized characteristic score (\ie, equals to 0 for $a_i$ with relevance score 0 and equals to 1 for $a_i$ with positive relevance score). 

More details of the three loss functions are given in supplementary materials. It is worth noting that our losses are derived from the underlying causal principle P2 in Eq.~\eqref{eq:5}, but not from the purpose of regressing to the ground-truth NDCG. The comparison will be given in Section~\ref{subsection:6_4}.

\subsection{Hidden Dictionary Learning} The aforementioned two implementations are discrete since they sample specific $a_i$ to approximate $u$. For better approximation, we propose learning to approximate the unobserved confounder $U$. As shown in Figure~\ref{fig:3_e}, we design a dictionary to model $U$. In practice, we design the dictionary as a $N \times d$ matrix $D_u$, where $N$ is manually set and $d$ is the hidden feature dimension. Note that given a sample $u$ and a answer candidate $a_c$, Eq.\eqref{eq:3} can be implemented as $\sum\nolimits_u P_u(a_c)P(u|H)$. Since the last layer of our network for answer prediction is a softmax layer: $P_u(a_c) = \text{softmax}(f_s(\bm{e_c},\bm{u},\bm{m}))$, where $\bm{e_c}$ is the embedding of candidate $a_c$, $\bm{u}$ is sampled from $\bm{D_u}$, $\bm{m}$ is the joint embedding for $\{Q,I,H\}$, and $f_s$ is a similarity computation function, the Eq.\eqref{eq:3} can be re-written as:
\begin{equation}
\centering
\label{eq:6_2}
P(A|\textit{do}(Q,H,I)):=\mathbb{E}_{\left[u|H\right]}\left[\text{softmax}(f_s(\bm{e_c},\bm{u},\bm{m}))\right].
\end{equation}
Since Eq.~\eqref{eq:6_2} needs expensive samplings for $\bm{u}$, we use NWGM approximation~\cite{xu2015show, srivastava2014dropout} to efficiently move the expectation into the softmax:
\begin{equation}
\centering
\label{eq:7}
\mathbb{E}_{[\!u|H\!]}[\text{softmax}(\!f_s\!(\bm{e_c},\!\bm{u},\!\bm{m}\!))] {\approx}
\text{softmax}(\mathbb{E}_{[\!u|H\!]}[\!f_s\!(\bm{e_c},\!\bm{u},\!\bm{m}\!)]).
\end{equation}
The details of the NWGM approximation can be found in supplementary materials. In this paper, we model $f_s(\bm{e_c},\bm{u},\bm{m}) = \bm{e_c}^T(\bm{u} + \bm{m})$. Thanks to the linear additive property of expectation calculation, we can use $\bm{e_c}^T(\mathbb{E}_{[u|H]}[\bm{D_u}] + \bm{m})$ to calculate $\mathbb{E}_{[u|H]}[\bm{e_c}^T(\bm{u} + \bm{m})]$. In practice, we use a dot-product attention to compute $\mathbb{E}_{[u|H]}[\bm{D_u}]$. Specifically, $\mathbb{E}_{[u|H]}[\bm{D_u}] = \text{softmax}(\bm{L}^T\bm{K})\odot \bm{D_u}$, where $\bm{L}=\bm{W_1}\bm{h}$, $\bm{K}=\bm{W_2D_u}$ and $\odot$ is element-wise product, $\bm{h}$ is the embedding of history $H$, and $\bm{W_1},\bm{W_2}$ are mapping matrices. The training details can be found in Section~\ref{subsection:6_3}. 
%%%%%%%%% 6 Experiments
\section{Experiments}
\label{section:6}
\subsection{Experimental Setup}
\noindent\textbf{Dataset.} Our proposed principles are evaluated on the recently released real-world dataset VisDial v1.0\iffalse\footnote{Suggest by the official~\cite{visdial2019}, results should be reported on v1.0 instead of v0.9}\fi. Specifically, the training set of VisDial v1.0 contains 123K images from the COCO dataset~\cite{lin2014microsoft} with a 10-round dialog for each image, resulting in 1.2M dialog rounds. The validation and test sets were collected from Flickr, with 2K and 8K COCO-like images respectively. The test set is further split into test-std and test-challenge splits, both with the number of 4K images that are hosted on the blind online evaluation server. Each dialog in the training and validation sets has 10 rounds, while the number in the test set is uniformly distributed from 1 to 10. For each dialog, a list of 100 answer candidates is given for evaluation. In the following, the results are reported on the validation and test-std set.

\noindent\textbf{Metrics.} As mentioned in Section~\ref{subsection:3_1}, NDCG is recommended by the official and accepted by the community. There are some other retrieval-based metrics like MRR (Mean Reciprocal Rank), where the ground-truth answer is generated by the single user. Note that the only answer may be easily influenced by the single user's preference (\ie., length). We argue that this may be the reason why the models with history shortcut achieve higher MRR, (\eg., due to the bias illustrated in Figure~\ref{fig:2}) and lower NDCG. Therefore, retrieval-based metrics are not consistent with NDCG. According to the mentioned reasons and space limitation, we only show the results on NDCG in the main paper. For completeness, the further discussion between NDCG and other retrieval-based metrics and the performance on all metrics will be given in the supplementary materials.

\subsection{Model Zoo} 
We report the performance of the following base models, including LF~\cite{das2017visual}, HCIAE~\cite{lu2017best}, CoAtt~\cite{wu2018you} and RvA~\cite{niu2019recursive}:

\noindent\textbf{LF}~\cite{das2017visual}. This naive base model has no attention module. We expand the model by adding some basic attention operations, including question-based history attention and question-history-based visual attention refinement.

\noindent\textbf{HCIAE}~\cite{lu2017best}. The model consists of question-based history attention and question-history-based visual attention.

\noindent\textbf{CoAtt}~\cite{wu2018you}. The model consists of question-based visual attention, image-question-based history attention, image-history-based question attention, and the final question-history-based visual attention.

\noindent\textbf{RvA}~\cite{niu2019recursive}. The model consists of question-based visual attention and history-based visual attention refinement. 

\subsection{Implementation Details}
\label{subsection:6_3}
\noindent\textbf{Pre-processing.} For language pre-processing, we followed the process introduced by~\cite{das2017visual}. First, we lowercased all the letters in sentences and converted digits to words and removed contractions. After that, we used Python NLTK toolkit to tokenize sentences into word lists, followed by padding or truncating captions, questions, and answers to the length of 40, 20 and 20, respectively. Then, we built a vocabulary of the tokens with the size of 11,322, including 11,319 words that occur at least 5 times in train v1.0 and 3 instruction tokens. We loaded the pre-trained word embeddings from GloVe~\cite{pennington2014glove} to initialize all word embeddings, which were shared in encoder and decoder, and we applied 2-layers LSTMs to encode word embedding and set their hidden state dimension to 512. For the visual feature, we used bottom-up-attention features~\cite{anderson2018bottom} given by the official~\cite{visdial2019}.

\noindent\textbf{Implementation of Principles.} For Principle 1 (P1), we eliminated the history feature in the final fused vector representation for all models, while kept other parts unchanged. For HCIAE~\cite{lu2017best} and CoAtt~\cite{wu2018you}, we also blocked the history guidance to the image. For Principle 2 (P2), we trained the models using the preference score, which can be counted from question type or given by the official (\ie, dense annotations in VisDial v1.0 training set). Specifically, for ``question type'', we first defined 55 types and marked answers occurred over 5 times as preferred answers, then used the preference to train our model by ($R_2$) loss proposed in Section~\ref{subsection:5_2}. ``Answer score sampling'' was directly used to fine-tune our pre-trained model by the proposed loss functions. For ``hidden dictionary'', we set a matrix for $N$ as 100 and $d$ as 512 to realize $D_u$. The dictionary is initialized with the features of top-100 popular answers, then trained by dense annotations with $R_3$ loss. More details can be found in supplementary materials. Note that the implementations following P1 and P2 are flexible.

\noindent \textbf{Training.} We used softmax cross-entropy loss to train the model with P1, and used Adam~\cite{DBLP:journals/corr/KingmaB14} with the learning rate of $4\times10^{-3}$ which decayed at epoch 5, 7, 9 with the decay rate of 0.4. The model was trained for 15 epochs totally. In addition, Dropout~\cite{srivastava2014dropout} was applied with ratio of 0.4 for RNN and 0.25 for fully connected layers. Other settings were set by default.
\begin{table}[t]
\centering
\scalebox{0.8}
{
\begin{tabular}{l | c | c | c  c  c  c | c }
\hline
\multirow{2}{*}{Model}&\multirow{2}{*}{baseline}&\multirow{2}{*}{QT}&
\multicolumn{4}{c|}{S}&{\multirow{2}{*}{D}} \\
\cline{4-7}
 & & & $R_0$ & $R_1$ & $R_2$ & $R_3$ &\\
\hline 
LF~\cite{das2017visual} & 57.21 &  58.97  & 67.82 & 71.27 & 72.04 & 72.36 & 72.65\\
LF +P1 & 61.88 &  62.87  & 69.47 & 72.16 & 72.85 & 73.42 & \textbf{73.63}\\
\hline
\end{tabular}
}
\vspace{2pt}
\caption{Performance (NDCG\%) comparison for the experiments of applying our principles on the validation set of VisDial v1.0. LF is the enhanced version as we mentioned. QT, S and D denote question type, answer score sampling, and hidden dictionary learning, respectively. $R_0$, $R_1$, $R_2$, $R_3$ denote regressive loss, weighted softmax loss, binary sigmoid loss ,and generalized ranking loss, respectively.}
\label{table:1}
\end{table}

\subsection{Quantitative Results}
\label{subsection:6_4}

Table~\ref{table:1} shows the results with different implementations in P2, \ie, question type, answer score sampling, and hidden dictionary learning. Overall, all of the implementations can improve the performances of base models. Specifically, the implementations of P2 can further boost performance by at most 11.75\% via hidden dictionary learning. Specifically, our designed loss functions based on Eq.~\eqref{eq:3} outperform the regressive score, which is implemented as Euclidean distance loss and denoted as $R_0$. The reason is that the regression fine-tune strategy is not a proper approximation for P2. We also find that the proposed ranking loss (\ie, $R_3$) performs best since it satisfies the ranking property of VisDial.

Note that our principles are model-agnostic. Table~\ref{table:2} shows the results about applying our principles on four different models (\ie, LF~\cite{das2017visual}, HCIAE~\cite{lu2017best}, CoAtt~\cite{wu2018you} and RvA~\cite{niu2019recursive}). In general, both of our principles can improve all the models in any ablative condition (\ie, P1, P2, P1+P2). Note that the effectiveness of P1 and P2 are additive, which means combining P1 and P2 performs the best.

We finally used the blind online test server to justify the effectiveness of our principles on the test-std split of VisDial v1.0. As shown in Table~\ref{table:3}, the top part contains the results of the baseline models implemented with our principles, while the bottom part represents the recent Visual Dialog Challenge 2019 leaderboard~\cite{visdial2019challenge}. We used the ensemble of the enhanced LF~\cite{das2017visual} to beat the Visual Dialog Challenge 2019 Winner (\ie, MReaL-BDAI), which can also be regarded as the implementations of P1 and P2. Promisingly, by applying our principles, we can promote all the baseline models to the top ranks on the leaderboard.

\begin{table}[t]
\centering
\scalebox{0.8}
{
\begin{tabular}{l | c  c  c  c }
\hline
Model & LF~\cite{das2017visual} & HCIAE~\cite{lu2017best} & CoAtt~\cite{wu2018you} & RvA~\cite{niu2019recursive} \\
\hline 
baseline & 57.21 & 56.98 & 56.46 & 56.74\\
\hline 
+P1 & 61.88 & 60.12 & 60.27 & 61.02 \\
+P2 & 72.65 & 71.50 & 71.41 & 71.44 \\
+P1+P2 & \textbf{73.63} & 71.99 & 71.87 & 72.88 \\
\hline
\end{tabular}
}
\vspace{1pt}
\caption{Performance (NDCG\%) of ablative studies on different models on VisDial v1.0 validation set. P2 indicates the most effective one (\ie, hidden dictionary learning) shown in Table~\ref{table:1}. Note that only applying P2 is implemented by the implementations in Section~\ref{section:5} with the history shortcut.}
\vspace{-2pt}
\label{table:2}
\end{table}
\begin{figure}[t]
\centering
\includegraphics[width=3.2in]{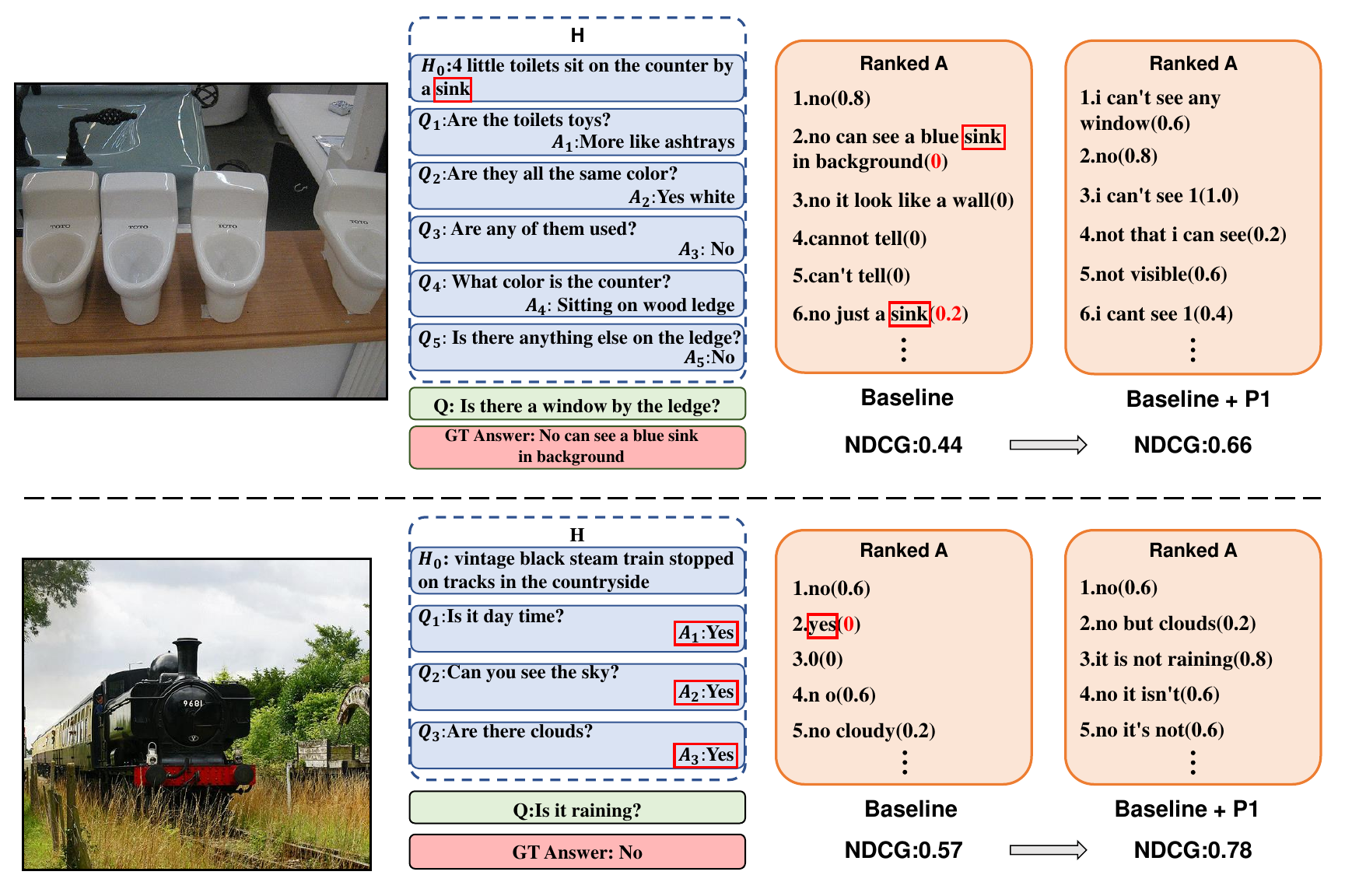}
\caption{Qualitative results of the baseline and baseline with P1 on the validation set of VisDial v1.0. The numbers in brackets in ranked $A$ denote relevance scores. Red boxes denote that the selected candidates of the baseline model influenced by the shortcut (\eg, word matching) from the dialog history. For the baseline with P1, it does not make such biased shortcut choices. More details can be found in Section~\ref{subsection:6_5}.}
\vspace{-2pt}
\label{fig:4}
\end{figure}

\begin{figure*}[t]
\centering
\includegraphics[width=170mm]{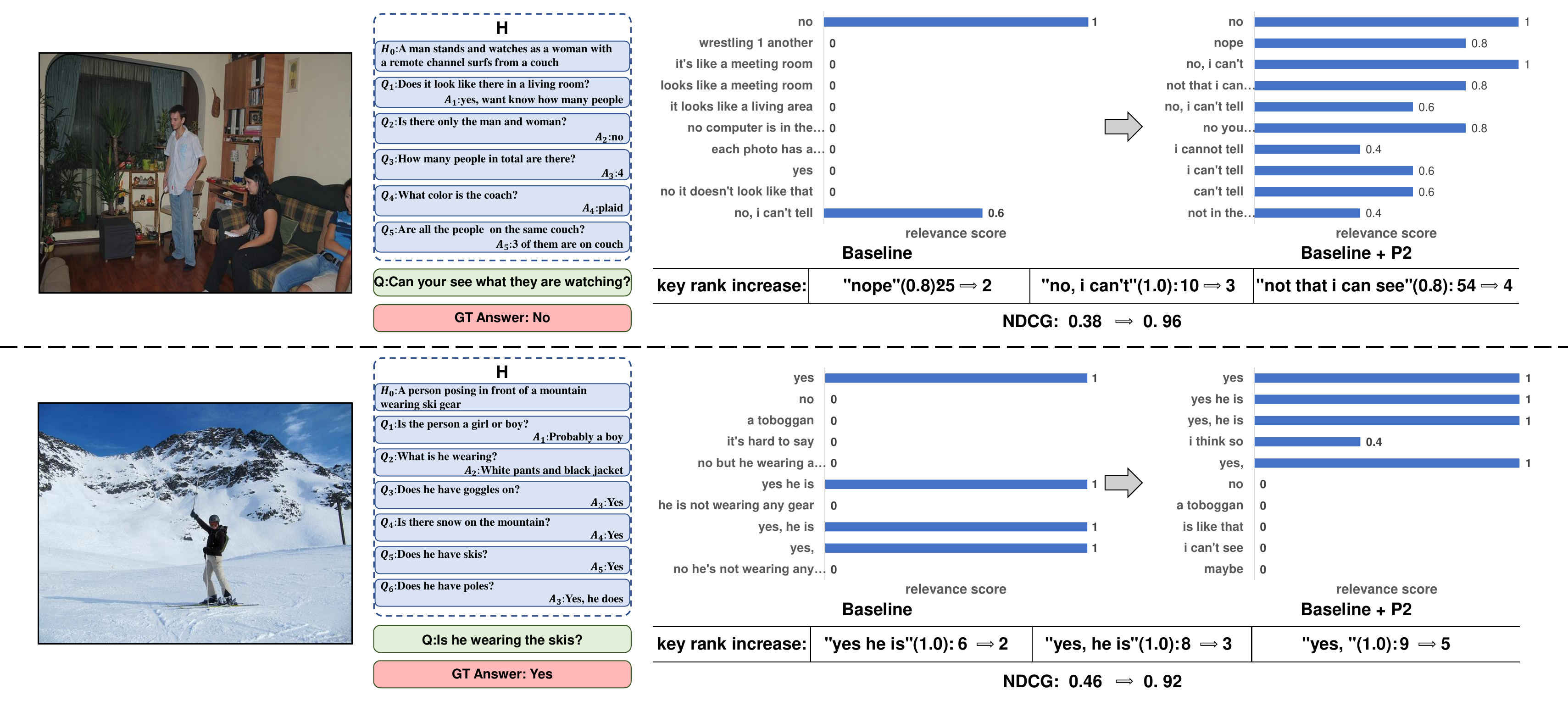}
\caption{Qualitative examples of the ranked candidates of baseline and baseline with P2. We also give some key rank changes for boosting NDCG performance by implementing P2. These examples are taken from the validation set of VisDial v1.0.}
\label{fig:5}
\vspace{-2pt}
\end{figure*}

\subsection{Qualitative Analysis} 
\label{subsection:6_5}
The qualitative results illustrated in Figure~\ref{fig:4} and Figure~\ref{fig:5} show the following advantages of our principles.

\noindent\textbf{History Bias Elimination.} After applying P1, the harmful patterns learned from history are relieved, like the answer-length bias shown in Figure~\ref{fig:2_a} as we mentioned. The example on the top of Figure~\ref{fig:4} shows the word-match bias in the baseline. From this example, we can observe that the word ``sink'' from history is literally unrelated to the current question, However, for the baseline model, some undesirable candidates (\ie, with low relevance score) containing the word ``sink'' can be found in the top-ranked answers due to the wrong direct history shortcut. To further confirm the conjecture, we counted the word-match cases of objective words (\eg, ``sink'' and ``counter'') on the validation set for the top-10 candidates of the ranked lists. The statistic indicates that P1 can decrease about 10\% word matching from history (from $\sim$5200 times of baseline to $\sim$4800 times using P1). The bottom example shows that, when the answer ``yes'' exists in history, the baseline model will tend to rank ``yes'' in a high place. However, it is opposite to the real answer ``no'' in some cases, which will lead to a lower NDCG. After applying P1, this problem can be effectively alleviated. To testify this conclusion, we further calculate the average ranks of ``yes'' for baseline and baseline with P1 in the above case (\ie, ``yes'' appears in history and the real answer is ``no''). We find that the average ranks are 4.82 for baseline and 6.63 for baseline with P1 respectively. The lower rank means that P1 relieves the ``yes'' shortcut in history. More examples of these biases can be found in supplementary materials.
\begin{table}[t]
\centering
\scalebox{0.8}
{
\begin{tabular}{ c | c | c}
\hline
 & Model & NDCG(\%) \\
\hline 
\multirow{6}{*}{Ours} & P1+P2 (More Ensemble) & 74.91\\
~ & LF+P1+P2 (Ensemble) & 74.19\\
 ~ & LF+P1+P2 (single) & 71.60\\
 ~ & RvA+P1+P2 (single) & 71.28\\
 ~ & CoAtt+P1+P2 (single) & 69.81\\
 ~ & HCIAE+P1+P2 (single) & 69.66\\
\hline
\multirow{5}{*}{Leaderboard} & VD-BERT(Ensemble)$^*$ & 75.13\\
 ~ & Tohuku-CV Lab (Ensemble)$^*$ & 74.88 \\
 ~ & MReaL-BDAI$^*$ & 74.02\\
 ~ & SFCU (Single)$^*$ & 72.80 \\
 ~ & FancyTalk (HeteroFM)$^*$ & 72.33\\
 ~ & Tohuku-CV Lab (Ensemble w/o ft)$^*$ & 66.53\\
\hline
\end{tabular}
}
\vspace{2pt}
\caption{Our results and comparisons to the recent Visual Dialog Challenge 2019 Leaderboard results on the test-std set of VisDial v1.0. Results are reported by the test server, ($^*$) denotes it is taken from~\cite{visdial2019challenge}. Note that the top five models in the Leaderboard use the dense fine-tune implementation illustrated in Section~\ref{subsection:5_2}.}
\vspace{-2pt}
\label{table:3}
\end{table}

\noindent\textbf{More Reasonable Ranking.} Figure~\ref{fig:5} shows that the baseline model only focuses on ground truth answers like ``no'' or ``yes'' and does not care about the rank of other answers with similar meaning like ``nope'' or ``yes, he is''. This does not match human's intuition because the candidates with similar semantics are still reasonable. This also leads the baseline model to a lower NDCG. As shown in Figure~\ref{fig:5} the model with P2 almost ranks all the suitable answers like ``yes, he is'', ``yes he is'' and ``I think so'' at top places together with the ground truth answer ``yes'', which significantly improves the NDCG performance.

%%%%%%%%% 7 Conclusions
\section{Conclusions}
In this paper, we proposed two causal principles for improving the VisDial task. They are model-agnostic, and thus can be applied in almost all the existing methods and bring major improvement. The principles are drawn from our in-depth causal analysis of the VisDial nature, which is however unfortunately overlooked by our community. For technical contributions, we offered some implementations on how to apply the principles into baseline models. We conducted extensive experiments on the official VisDial dataset and the online evaluation servers. Promising results demonstrate the effectiveness of the two principles. As moving forward, we will stick to our causal thinking to discover other potential causalities hidden in embodied Q\&A and conversational visual dialog tasks.\\

\noindent\textbf{Acknowledgement} This work was partially supported by the National Natural Science Foundation of China (61573363 and 61832017), the Fundamental Research Funds for the Central Universities and the Research Funds of Renmin University of China (15XNLQ01), and NTU-Alibaba JRI. We would like to thank the anonymous reviewers for their constructive comments.

%%%%%%%%% 8 Acknowledgement
% \input{sections/8_Acknowledgement.tex}
{\small
\bibliographystyle{ieee_fullname}
\bibliography{egbib}
}
\clearpage

% \includepdf[pages={1}]{supp.pdf}
% \includepdf[pages={2}]{supp.pdf}
% \includepdf[pages={3}]{supp.pdf}
% \includepdf[pages={4}]{supp.pdf}
% \includepdf[pages={5}]{supp.pdf}

\end{document}